\documentclass[a4paper,10pt,conference]{ieeeconf}
\usepackage{graphicx} 
\usepackage{cite}
\usepackage{mathptmx}
\usepackage{times}
\usepackage{url}
\usepackage{amsfonts, amsmath}
\usepackage{bm}
\usepackage{amssymb}
\usepackage{interval}
\usepackage{tabularx}
\usepackage{multirow}
\usepackage{textcomp}
\usepackage{xcolor}
\usepackage{comment}
\title{An Open Source Computer Vision and Machine Learning Framework for Affordable Life Science Robotic Automation}
\author{Zachary Logan, Andrew Dudash, and Daniel Negr\'on}

\begin{document}

\maketitle
\thispagestyle{empty}
\pagestyle{empty}

\begin{abstract}
	We present an open-source robotic framework that integrates computer vision and machine learning based inverse kinematics to enable low-cost laboratory automation tasks such as colony picking and liquid handling. The system uses a custom trained U-net model for semantic segmentation of microbial cultures, combined with Mixture Density Network for predicating joint angles of a simple 5-DOF robot arm. We evaluated the framework using a modified robot arm, upgraded with a custom liquid handling end-effector. Experimental results demonstrate the framework's feasibility for precise, repeatable operations, with mean positional error below \(1\) mm and joint angle prediction errors below \(4\) degrees and colony detection capabilities with IoU score of \(0.537\) and Dice coefficient of \(0.596\).
\end{abstract}

\section{Introduction}
Automation of repetitive tasks has become a cornerstone of modern life sciences and has enabled laboratories to process larger volumes of samples~\cite{tacker2014,choi2018} with improved efficiency~\cite{da2016, hawker2000}, reproducibility~\cite{klevebring2009, holland2020}, and safety~\cite{movsisyan2016, caragher2017}. It has also reduced the workload experienced by lab technicians~\cite{swangnetr2018} and the cost~\cite{archetti2017} and number of man hours~\cite{holland2020} required to complete a protocol. However, many academic laboratories still maintain a large volume of research protocols that rely on manual operations~\cite{holland2020}. While many commercially available automation platforms such as those offered by Molecular Devices, Tecan, Hamilton, and TAP Biosystems have successfully automated many aspects of protocol pipelines, their effectiveness is directly correlated with their cost. These automation solutions can cost up to one million dollars~\cite{storrs2013}, making them out of reach for many academic laboratories. 

Other more affordable commercial platforms such as the Opentrons OT-2 and the Andrews+ liquid handling robot demonstrate the potential for more accessible laboratory automation solutions. These systems provide well designed, easy-to-use, user-friendly interfaces, and programmable protocols for tasks like pipetting, sample mixing, and plate handling at a fraction of the price of traditional liquid handling robots. However, such solutions are also often constrained by proprietary hardware, limited flexibility for reconfiguration, and additional costs for closed-source software and patented consumables, resulting in vendor lock-in~\cite{holland2020,huggett2009,moutsatsou2019}. Some examples include the pipette tips for Opentron's and Tecan's liquid dispensing system and spin kits for Qiagens Qiacube system.  

To address these issues, we seek to develop a framework for an autonomous system to perform various monotonous tasks using open-source software and low-cost hardware to simplify and facilitate the adoption of automation in early-stage lab procedures, thereby reducing lab staff workload. \textcolor{black}{In this paper, we focus on developing a framework that decouples perception, kinematic modeling, and robot hardware, allowing academic and other small laboratories to automate repetitive tasks using a simple machine learning (ML) based computer vision (CV) model combined with an ML based inverse kinematic (IK) solver to integrate existing or custom manipulators.}
\textcolor{black}{\begin{itemize}
	\item We construct a low-cost open-source experimental platform using commodity hardware to evaluate an ML based CV and IK framework for vision guided microbial colony sampling. 
	\item We build a perception system that only requires a few hundred sample images to effectively detect microbial colonies.
	\item We experimentally verify the efficacy of colony detection after 20 trials to quantify segmentation accuracy.
	\item We experimentally validate the end-to-end vision-to-action framework by performing randomized colony picking tasks.
\end{itemize}}

\section{Related Work}

Automated colony picking and sample identification through CV has long been an area of interest in the life sciences field. Early implementations relied on classical CV techniques such as edge detection, local thresholding, contour detection, and size measurements to identify colonies on plates~\cite{jones1992,uber1990,hansen1993}. However, these approaches often require manual image adjustment before drawing bounding boxes to mark regions of interest that contain colonies~\cite{wang2011}. The CV analysis is then applied to determine valid colony locations within the image and determine which can be picked. In addition, these approaches often struggle with lighting conditions optimized for colony phenotype and agar composition, requiring additional parameter fine-tuning that we seek to avoid in our detection system~\cite{sandle2020}.  

Further processing requires transferring the samples to a new location or type of container that depends on the downstream protocol. For instance, an isolated colony with a desired trait is streaked onto a new dish or transferred into a series of test tubes or well plate positions. Currently, we focus on a proof-on-concept for \textcolor{black}{moving colonies bewtween dishes} with the goal of mixing samples with reagents to prepare them for molecular diagnostics research, such as multiplexed polymerase chain reaction (PCR), loop-mediated isothermal amplification (LAMP), and antigen detection kit development. Accordingly, the ideal solution is a general-purpose liquid handling robot (GPLHR) that facilitates integration with \textcolor{black}{lab supplies} (e.g. pipette tips), hardware, and software by non-technical end-users~\cite{li2015}. For example, recent progress towards a GPLHR by Knobbe et al. developed a pipetting system adhering to ISO 8655 standards within a modular hardware and software architecture~\cite{knobbe2022}. Also, the widespread availability of cheap microcontrollers, 3D-printing, laser-cutting, and LEGO® have lowered the barrier to entry for those laboratories able to draw from inexpensive DIY projects and open-source robotics~\cite{gerber2017,li2022}. 

To perform \textcolor{black}{colony sampling} using a robotic manipulator, one will have to obtain a solution to the IK problem. \textcolor{black}{There are multiple IK solutions: numerical methods such as pseudoinverse~\cite{whitney1969}, cyclic coordinate descent~\cite{wang1991}, Jacobian Transpose~\cite{balestrino1984, wolovich1984}, and Levenberg-Marquardt damped least squares~\cite{wampler1988, nakamura1986}. While numerical methods can generalize to a larger number of robot manipulators, many have poor performance near singularities and are sensitive to the initial guess~\cite{buss2004}. Moreover, numerical methods often assume or require perfect encoder or position feedback, which is not always achievable when using lower cost hardware.} 

\textcolor{black}{Obtaining one is a non-trivial task that can be further complicated by robot manipulators containing redundant joints, flexible joints, or low repeatability. Traditional approaches to solving the IK problem relied on analytical solutions; however, it is not always possible to obtain an analytical IK solution, which work best for robot manipulators with simple kinematic chains, such as those containing only prismatic and revolute joints with a spherical wrist~\cite{pieper1969}. Pieper\textcolor{black}{~\cite{pieper1969}} derived closed-form solutions for specific robot manipulator configurations; however, these solutions do not generalize to custom configurations.}

To bypass the challenges of solving the IK problem, recent work has explored using ML techniques to approximate the solution directly, including artificial neural networks~\cite{duka2014}, convolutional neural networks~\cite{levine2018}, modular neural networks~\cite{oyama2001}, and Widrow-Hoff neural networks~\cite{ramdane2002}. ML approaches have been used to successfully implement neural networks to solve the IK problem for redundant manipulators~\cite{vu2023} and integrate collision avoidance~\cite{tenhumberg2023}. Other studies have also used ML approaches to directly learn the mapping required for visual servoing~\cite{levine2018}. However, many solutions struggle with the ambiguity of finding multiple solutions to the IK problem. One remedy is to map the entire solution space using a mixture density network, which is capable of greatly improving position errors~\cite{bishop1994}.

\section{Methods}

\subsection*{Computer Vision}
\textcolor{black}{To localize microbial colonies within the Petri dish, we opted to use a semantic segmentation classifier with a U-net architecture~\cite{ronneberger2015} due to its ability to provide pixel-level representations of the colonies. This allows us to maintain the full spatial geometry of the colonies, which is critical for downstream sampling where the sampling location must be within the biological media. Further, the U-net semantic segmentation classifier is considered as the gold standard technique in medical imaging due to its ability to obtain accurate results with only a small amount of data~\cite{safarov2025}.} 

In our setup, the neural networks comprising the U-net model accepted 3-channel RGB images as inputs and output a segmentation mask corresponding to identified colonies. In this experiment, \textcolor{black}{we were interested in identifying generic colonies with the ability to further integrate species specific and phenotype-based detection.}


\subsubsection*{Dataset Generation}
The training dataset consisted of 28 labeled images, which was expanded using a combination of image augmentation techniques via the OpenCV library~\cite{bradski2000}, such as rotations, to a total of 448 images. The original 28 images were labeled using the Labelme software~\cite{wada2025}. During training, the final dataset was randomly split into two subsets with 80\% of the images for training and 20\% of the images for validation. 

\subsubsection*{Model Training}
Training was conducted on an Nvidia Ada-series RTX 2000 GPU with 8 GB of VRAM. The model was trained over 80 epochs with a learning rate of \(5\times{10}^{-4} \) and batch size of 8 images. The model was optimized using the \textcolor{black}{AdamW optimizer~\cite{loshchilov2017decoupled}} using the binary cross-entropy with logits loss. Performance was evaluated on the validation set at the end of each epoch. 

\subsection*{IK Model}

\subsubsection*{Dataset Generation}
Ground truth data was generated using the Robotics Toolbox Python Library developed by Peter Corke~\cite{corke2021}. First, using the ERobot function, we created a representation of the robot using its URDF definition and masked any joints not involved in the position control (e.g. the syringe motor). Joint angle configurations were sampled uniformly within the robot's defined joint limits using Equation~\ref{eq:joint_sampling}. Then for each sampled joint configuration \( \mathbf{q}_{i} \), and end-effector pose \((\mathbf{x}_{i},~\mathbf{\theta}_{i}) \), the forward kinematics function was used to obtain the Cartesian position of the end-effector. As the model was designed to work with a CV algorithm, we only stored the translation components of the pose. We repeated this process until we had 10,000 \( [\mathbf{x}_{i},~\mathbf{q}_{i}] \).
\begin{equation} \label{eq:joint_sampling}
	q_{i} \sim \mathcal{U}(q_{min}, q_{max})
\end{equation}
\subsubsection*{Mixture Density Network Setup}
\textcolor{black}{To support a flexible and robot-\textcolor{black}{agnostic} framework, we implemented a data driven IK model that learns the relationship between the joint space and end-effector space for any robot system directly from the data. This design allows our framework to be applied to any robot platform\textcolor{black}{, regardless of size,} without requiring kinematic derivations or manual reconfiguration. As IK solutions naturally contain multiple valid joint configurations for a given end-effector position, we chose to represent this structure by implementing a Mixture Density Network (MDN). MDNs predict the conditional distribution \(p(\mathbf{q},~\mathbf{x}) \) as a weighted sum of \textbf{K} Gaussians.} Our MDN model took as inputs the three translation components of the end-effectors pose and outputs the parameters of a Gaussian model over the joint space of the robot using \(\mathbf{K} = 5\) Gaussians. Internally the model used a feedforward neural network architecture with three hidden layers consisting of 128 nodes each and activated using the Sigmoid-weighted Linear Unit (SiLU) function. The final layer breaks into three branches for predicting the mixture weights using SoftMax, the joint angle means, and the joint angle standard deviations using \textcolor{black}{softplus} for each Gaussian component.  

The MDN model was trained using the negative log likelihood loss function defined in Equation~\ref{eq:mdn_loss} of the true joint configuration under the predicted \textcolor{black}{Gaussian} mixture with the Adam Optimizer\textcolor{black}{~\cite{kingma2014adam}}. During training we used an exponential learning rate schedule with an initial rate of \(1.0\times{10}^{-2}\) that decayed every 100 epochs by a factor of \(0.90\). 
\begin{equation} \label{eq:mdn_loss}
	Loss = -\log{\sum_{k=1}^{k} \pi_{k}\mathcal{N}(\mathbf{q}_{true} | \mu_{k},~\sigma_{k}^{2})}
\end{equation}

\subsubsection*{Training Procedure}
Training was conducted on a Nvidia Ada-series RTX 2000 GPU with 8 GB of VRAM. The model was trained for 1000 epochs with a batch size of 256 using our pre-generated dataset. The MDN loss was backpropagated directly and there were no online computations for forward kinematics. During model evaluation the final joint angles were obtained by selecting the highest mode from the mixture distribution and using its mean as the predicted joint angle configuration. Model accuracy was evaluated to be within 2 degrees, by comparing the predicted joint configuration against a set of joint configurations used to generate a known manually coded demonstration trajectory.  

\subsection*{Prototype Test System}

\begin{figure}[!ht]
	\centering
	\includegraphics[width=\columnwidth]{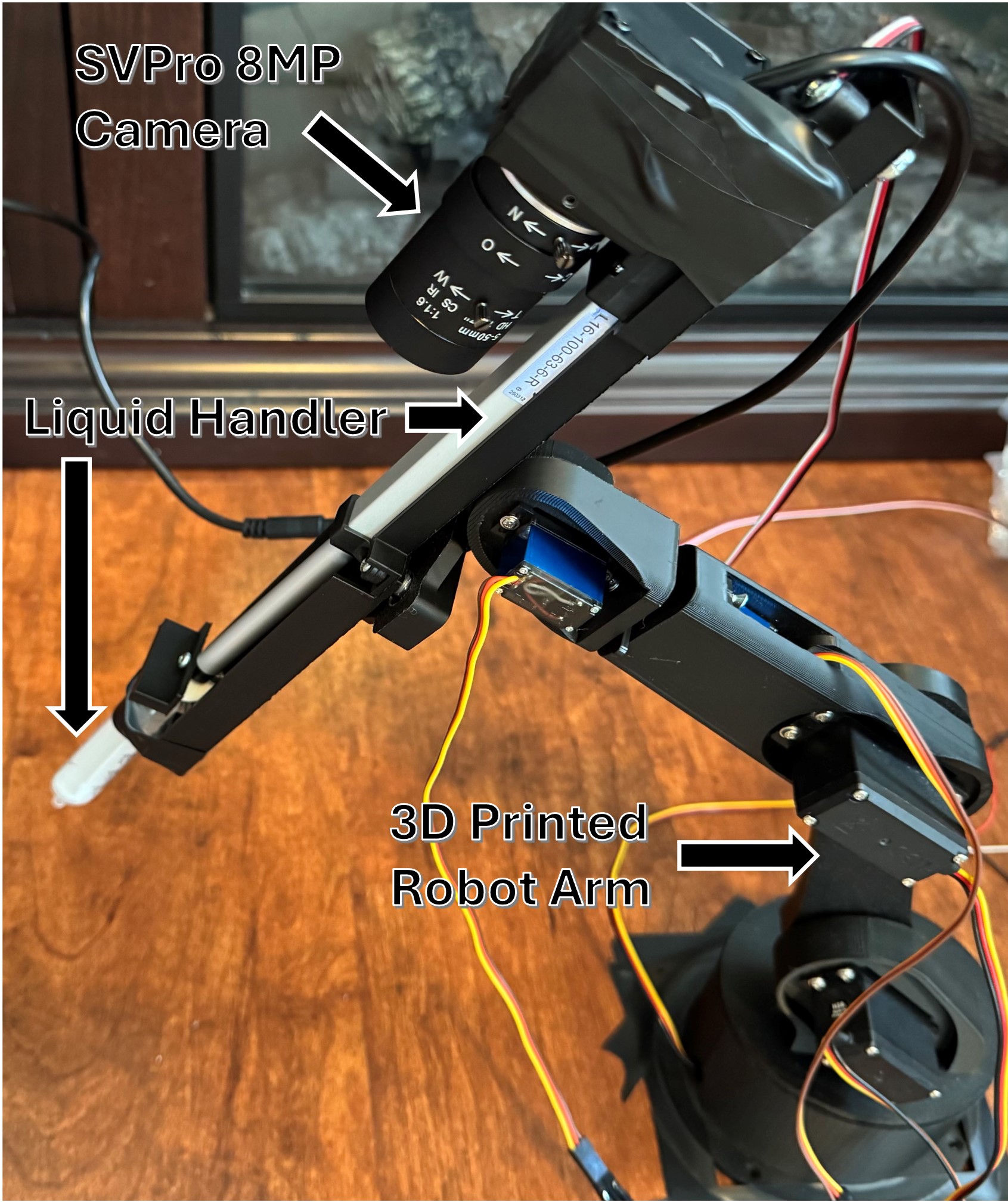}
	\caption{To automate the colony detection and sampling process, we developed a low-cost and open source based robotic platform integrating a modified 5-DOF robot based on the design in~\cite{dejan2023} chosen for its ease of assembly, readily available parts and affordability. The arm was fitted with upgraded Hitec D645MW and Towerpro MG92B servos for improved precision. At the end-effector, we attached an open-source digital \textcolor{black}{pipette} adapted from Yoshikawa et al.~\cite{yoshikawa2023} with a 10 mL syringe, enabling the precise aspiration and dispenstion of liquids and \textcolor{black}{a} high resolution SVPRO 8 MP USB camera. To control the servo motors we used an Arduino Uno microcontroller~\cite{arduino_uno} because of its popularity in robots, expansive documentation, open design, and low cost.}
	\label{fig:robotic_arm}
	\vspace{-4pt}
\end{figure}

\paragraph*{Robot Arm}
\textcolor{black}{To test how our computer vision and IK framework would perform on an example custom manipulator,} we modified a robotic manipulator (Figure~\ref{fig:robotic_arm}) based on an open-source 3D-print design from HowToMechatronics~\cite{dejan2023}. The modified  manipulator has 4 degrees of freedom configured with shoulder joints, an elbow joint, a wrist pitch joint and an end-effector. To increase the precision of the manipulator, we opted to use Hitec D645MW servos in place of the Towerpro MG996R and Towerpro MG92B servos in place of the SG90 micro servos as they offered better torque stability, durability, and internal feedback resolution. Supplementary Table S1~\cite{logan2025supplemental} lists the bill of materials and costs totaling under \$500.

\paragraph*{Digital Pipette}
The default end-effector was modified to include a liquid handling tool based on the open-source digital pipette design presented by Yoshikawa et al.~\cite{yoshikawa2023}. This design integrates an Actuonix L16-100-63-6-R precision linear actuator that drives a 10 mL plastic syringe, allowing for automated aspiration and dispensing of liquids. The actuator was mounted along the forearm of the original gripper as shown in Figure~\ref{fig:robotic_arm} and controlled in parallel with the other joints. 

\paragraph*{Vision System}
A high-resolution USB camera (SVPro 8 MP, 5-50 mm varifocal lens) was attached to the rear of the liquid handler to provide RGB images for the real-time semantic segmentation. The camera captured images at a resolution of 480p and at a rate of 30 fps. 

\begin{figure}[!ht]
	\centering
	\includegraphics[width=0.85\columnwidth]{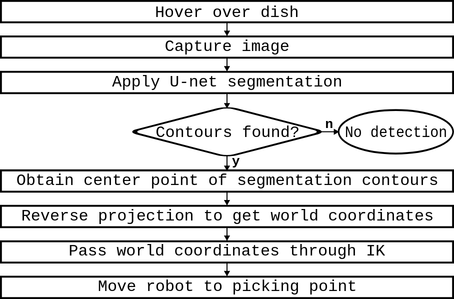}
	\caption{The Flow chart illustrates the control architecture for the automated colony picking system. The robot first hovers over the Petri dish and captures an image using the camera. The image is passed through the trained U-net segmentation model, if any colonies are detected, the center points of the contours bounding the colonies are extracted and converted from pixel coordinates to world coordinates. These coordinates are passed through the MDN-based IK model to compute the required joint angles. Lastly the robot arm moves to the picking point to acquire the sample. If no colonies are detected, the system outputs a message stating that nothing was found and all samples have been collected.}
	\label{fig:control_flowchart}
	\vspace{-16pt}
\end{figure}

\paragraph*{Control Architecture}
The entire system was controlled using a 4 GB Jetson Nano running Ubuntu 20.04 and an Arduino Uno microcontroller. The Jetson Nano was interfaced with the Arduino Uno using serial communication and the Robotic Operating System (ROS) Foxy Fitzroy distribution~\cite{macenski2022}. The prototype system used a 2-stage control process, outlined in Figure~\ref{fig:control_flowchart}: an image captured by the USB camera is processed by a trained segmentation model (U-net) to detect colonies contained within a Petri dish located within the workspace. The pixel coordinates are then transformed into the world frame coordinates using a reverse projection based on the pinhole camera model. The world frame coordinates are then passed to the Mixed Density Network (MDN) IK model to infer the joint angles needed to reach the desired Cartesian position. The computed joint angles are sent over the serial connection using a ROS 2 Foxy node to the Arduino microcontroller~\cite{arduino_uno}, which then executes smooth position control of each servo using the Servo library for precise movements.

\subsection*{\textcolor{black}{Testing Autonmous Accuracy}}
To determine whether our framework can effectively construct an autonomous colony sampling and processing robot capable of replacing or augmenting parts of the human workflow, we devised a set of experiments to test the integrated systems' accuracy, runtime\textcolor{black}{,} and practical reliability. These experiments were designed to evaluate the performance of each individual component, including the CV segmentation model, the IK prediction model, and the end-to-end pick-and-place cycle under realistic conditions. To assess whether the system meets the baseline for practical laboratory use, we recorded key metrics such as the segmentation quality, positional angular accuracy, and total task duration. 

\begin{figure}
		\centering
	\includegraphics[width=\columnwidth]{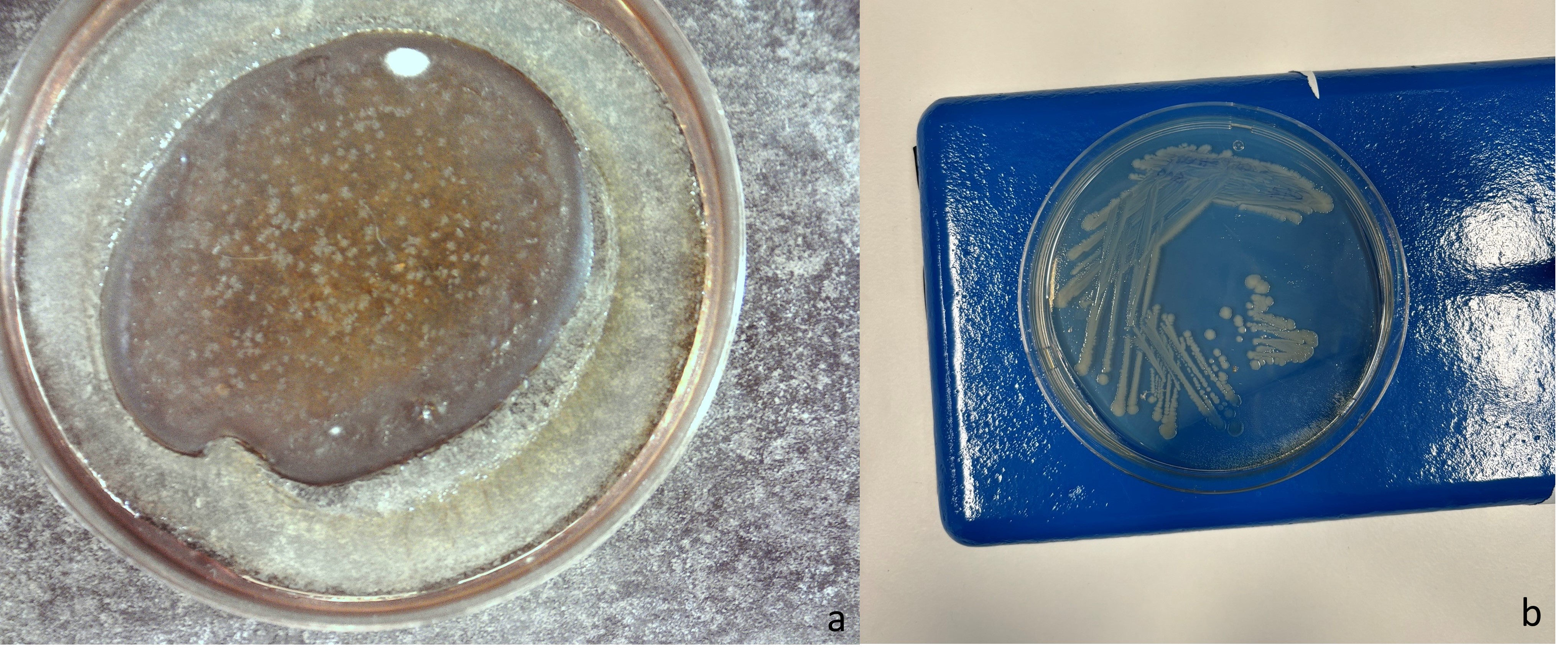}
	\caption{Representative samples from the agar plate image dataset used for training and testing the U-net segmentation model. a) A homemade agar plate using beef bouillon, sugar and gelatin, which produced slightly clear areas along the outer edge due to evaporation of some of the water used in production. b) A standard laboratory prepared agar plate using conventional techniques and materials, resulting in much cleaner and richer bacterial colony growth.}
	\label{fig:agar_samples}
	\vspace{-14pt}
\end{figure}

\paragraph*{Experiment 1: Image Segmentation Accuracy}
In this experiment, the objective was to quantify the segmentation accuracy of the custom U-net model under realistic imaging \textcolor{black}{conditions in a professional academic laboratory and in a DIY Lab, as shown in Figure~\ref{fig:agar_samples}.} A new test dataset was generated from images captured using the same SVPRO 8 Megapixel camera, example images shown in Figure~\ref{fig:agar_samples}. Each image was processed through the trained U-net model to generate a binary mask of microbial culture which was then compared to the manually annotated ground truth mask. For every image, we computed the Intersection over Union (IoU) and the Dice Coefficient to evaluate segmentation performance. Additionally, the inference time for each frame was recorded to verify the feasibility of real-time execution. 

\paragraph*{Experiment 2: IK Prediction Accuracy}
In this experiment, we assessed the accuracy of the MDN IK model by comparing its predicted joint configurations to ground truth joint angles generated with the Robotics Toolbox Python Library. A set of 15 Cartesian target positions obtained from a sample liquid transfer demonstration were defined as test inputs. The Cartesian positions were passed through the trained MDN to obtain predicted joint angles. The absolute error for each joint was calculated by comparing the predicted joint angle with the ground truth joint angle from the demonstration. When multiple Gaussian modes were available, the mode with the highest mixing coefficient was selected as the final prediction to reflect the most probable solution. We also recorded the time required to complete each inference to estimate the computational overhead. 

\paragraph*{Experiment 3: Full Experimental Pipeline}
In the last experiment, we analyzed the full end-to-end performance of the framework to \textcolor{black}{evaluate} real-world integration. For this experiment, \textcolor{black}{the agar plates containing isolated colonies were} placed at a known position within the robot's workspace. This position would be passed through the trained MDN model to obtain the joint angles required to position the camera above the Petri dish. A single image of the Petri dish is then captured by the camera and segmented using the trained U-net model. The pixel coordinates from the segmented image are transformed into world coordinates and passed through the trained MDN model to obtain the required joint angles to position the syringe above the culture. The predicted joint angles are then sent over serial to the Arduino Uno microcontroller, which moved the robot to the final reach location. \textcolor{black}{During this process, we simulated the final sampling operation as we shortened the length of the syringe causing the to robot hover over the isolated colony just above the surface of the agar. This simulated sampling was done to ensure consistency between trials by allowing repeated use of the same agar plates.} If the final position of the robot arm would result in a failed sample extraction (not overtop of the culture), we then \textcolor{black}{manually measured the distance, in millimeters, between the robot's position and the closest possible position that would have resulted in a successful sample extraction, in order to determine the to determine propagated positioning error of the pipeline.}


\section{Results}

\subsection*{Experiment 1: Image Segmentation Accuracy}
The developed U-net-based semantic segmentation model achieved an intersection over Union (IoU) score of \(0.537\) and a Dice coefficient of \(0.596\) on our test dataset comprised of 64 images. Although the pixel-wise classification accuracy was relatively high at 99\%, the moderate IoU and Dice scores indicate that while the model reliably distinguishes the general area and location of colonies, it struggles with precisely delineating the full extent of the colony boundaries. The cause of the struggle with colony boundaries likely results from our highly limited dataset, that was expanded using image augmentation techniques, and having the model be overfit to the training data. Another explanation for the more moderate scores could also be reflected from the manual labeling process of the training images not precisely matching the actual boundaries of the colonies contained within. While this performance may not be suitable for high-risk situations such as medical processing, we believe this performance is more than sufficient as proof that this method is viable for use in automating a small-scale lab. The inference time of our U-net model was \(49.3\) milliseconds, which is more than sufficient for supporting real time or near real time systems running on embedded hardware.  

\begin{figure}
		\centering
	\includegraphics[width=\columnwidth]{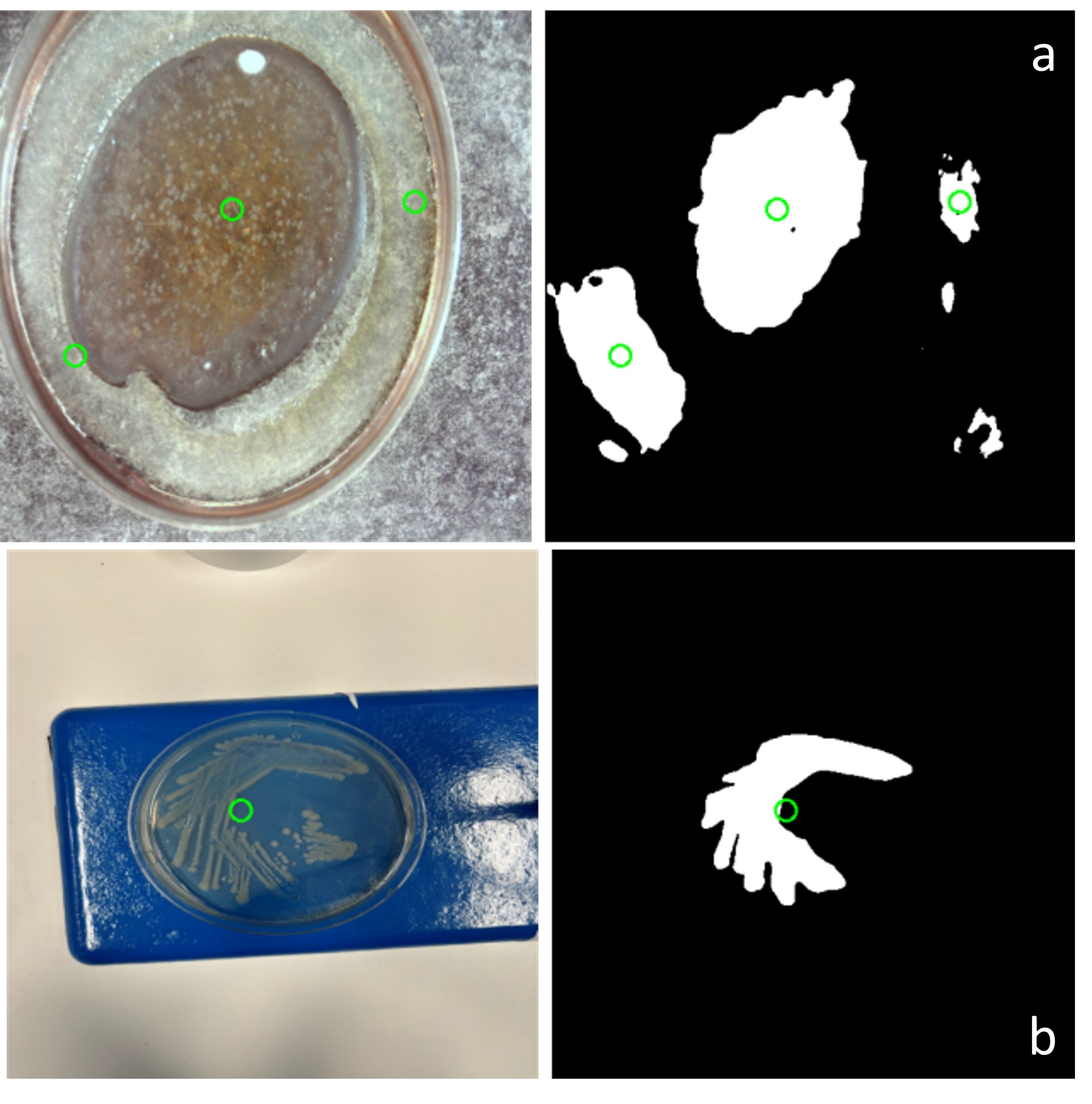}
	\caption{Example outputs from the semantic segmentation model on two agar plate images. (Left): Original Input image of an agar plate; (Right) Corresponding predicted segmentation mask. The green circles indicate the center point of the detected colonies and indicate the input used for the downstream IK calculations. a). Homemade agar plate prepared using beef bouillon, sugar and gelatin. b). Standard laboratory prepared agar plate.}
	\label{fig:example_segmentation_output}
	\vspace{-16pt}
\end{figure}

Example segmentation outputs including the original image, the predicted mask and an overlay of green circles to denote sampling locations, are shown in Figure~\ref{fig:example_segmentation_output}. The results illustrate that the model consistently identifies the approximate center point of \textcolor{black}{an isolated colony} with reasonable accuracy. 

\subsection*{Experiment 2: IK Prediction Accuracy}
The MDN-based IK model achieved an average worst-case end-effector position error of \(\pm 1\) millimeter and an angular joint error of \(\pm 4\) degrees across our test dataset of predefined Cartesian coordinates. On the Jetson Nano, the model's average inference time was \(0.93\) milliseconds, demonstrating that the MDN approach is highly suitable for real-time operation on highly resource constrained hardware and embedded systems. \textcolor{black}{The positional accuracy of \(\pm 1\) millimeter was within the practical capabilities for position repeatability of a low-cost or imprecise manipulator, where mechanical compliance, actuator resolution, and calibration error dominate overall accuracy. This result validates the MDN's ability to learn an effective mapping from end-effector Cartesian coordinates to joint configurations that is suitable for deployments on such manipulators.} However, our implementation has a higher-than-desirable joint angle error, and is a clear sign of ambiguity in the possible solutions to any IK problem: for any given end-effector pose, there exist multiple joint configurations that can achieve it. We believe that the higher joint angle error is a direct result of multiple joint configurations being valid for some of the test positions. Given that our robot arm is using hobby grade remote control vehicle servos with limited precision and some mechanical backlash, the practical effect of a few degrees of joint error would be minimal at the end-effector. Overall, these results indicate that the MDN approach is a promising and efficient alternative to conventional IK methods and will be easier for an end user to work with as training would only require a URDF model of the robot.

\subsection*{Experiment 3: Full Experimental Pipeline}
For the full integration of the CV image segmentation, the IK prediction and the control pipeline, our system had an average runtime of \(48.6\) seconds for completing a single colony detection and sampling operation. The runtime includes image capture, U-net image segmentation, coordinate transformation, IK solution and physical actuation. It is important to note that this runtime may be inflated due to a hardware limitation on the Jetson Nano which was powered using a 5-volt and 2-amp power supply during testing. This resulted in performance throttling due to overcurrent warnings and protections, which reduced the computational performance during the image processing and model inference. Increasing to a proper amperage power supply may greatly reduce the runtime.


During practical testing, the integrated system successfully detected colonies and guided the robot arm into the correct location for sampling with a similar deviation as found by the IK model, resulting in consistent sampling of the desired colony. However, we observed a consistent offset between the joint angles obtained by the IK model and the actual joint angle commands required by the Arduino to reach the same point. This mismatch can be caused by any number of factors, including a mismatch of joint axes due to differences in attaching the different parts on the servo or just differing reference frames. Specifically in our case, we found that mismatch was directly caused by a difference in definition between the frames defined in the URDF robot model used to train the IK network compared to the physical servo control scheme. For example, some of joints had an axis that was flipped and another had a shifted zero degree reference position, meaning instead of 0-180 degrees expected by the Arduino the IK had a positional range of 45-225 degrees. As a result, we applied a manually determined correction factor to the value of the joint angles to ensure the output of the MDN model was properly aligned with the values expected by the Arduino. These manual adjustments did not impede the system's ability to reach the desired picking points; however, they highlighted a key challenge for automating generic manipulators that can be controlled by any number of microcontrollers. \textcolor{black}{For ease of use, a small calibration program could be created that attempts to move the arm to a pre-set location and then prompts the user to measure and input the offset.}

\textcolor{black}{In our framework, we \textcolor{black}{derived} sampling locations based on the geometric center of segmented colony contours, which we noticed may not always align with the most desirable sampling location. This behavior was most common in colonies with highly irregular or non-convex morphologies such as in Figure~\ref{fig:example_segmentation_output}b. This limitation is derived from the post-processing choice for determing the sampling location within the mask and not the segmentation model itself, meaning that the overall framework can still effectively find \textcolor{black}{generic colonies accurately.} Futhermore because U-net produces a dense pixel-level representation of the colony and its geometry, our framework supports the integration of alternative target selection methods, without requiring the retraining of the perception model. As a result, our framework retains the flexibility required for the end user to implement more advanced biologically informed sampling strategies.}

\textcolor{black}{Overall, we have shown that this framework is a promising method for automating parts of manual lab procedures. Our framework allows for labs to have an autonomous system that is cheaper than current commercial options, is generic and open-source capable and can conceivable be trained and used by lab staff without any specialized training.}

\section{Conclusion}
In this paper, we designed, developed, and demonstrated an open source based, low-cost, prototype robotic framework to automate \textcolor{black}{microbial} bacteria colony detection and sampling tasks for small life sciences laboratories. We combined a U-net-based CV segmentation model trained on a custom dataset for colony segmentation with an MDN model for IK prediction and showed that this framework can accurately identify and reach \textcolor{black}{microbial} bacterial colonies on agar plates using 3D printed 5-DOF robot arm. Experimental results demonstrated that the segmentation model achieved an average IoU score of \(0.537\), an average Dice score of \(0.596\), a pixel-wise accuracy of 99\% and an inference time of \(49.3\) milliseconds; while the IK model achieved positional accuracy of \(\pm 1\) millimeter, angular accuracy of \(\pm 4\) degrees and an inference time of \(0.93\) milliseconds. Full pipeline tests confirmed that the system can execute the autonomous pick-and-place task with a run time of \(48.6\) seconds, demonstrating the feasibility of using affordable hardware and open-source methods to replicate parts of manual lab procedures.

While this prototype framework showed promising results, in future testing we will expand training and validation set sizes to enable more rigorous evaluation and comparisons to baseline systems. Future studies may focus on investigating different ways to improve integration speed and further reduce the effort required by the end user. One such investigation may examine the possibility of implementing large language models (LLMs) for automatic segmentation, automatic labeling of new training images collected during the run time and refinement of segmentation model, automatic creation and refinement of IK models based on performance during run time. This implementation would primarily focus on ways to use an LLM to reduce the minimum amount of domain knowledge required by the end user and allow them to provide performance feedback using natural language. \textcolor{black}{Future work will also investigate the addition of an online calibration program to allow users a simple method to correct for any offsets in the system and the integration of various strategies for determining the desired sampling location based on more advanced criteria and domain knowledge of the type of microbe.} \textcolor{black}{Future studies will examine a comparison between} using a similar framework on a gantry mechanism in place of the robot arm or in combination with a robot arm, to reduce the complexity of IK solutions while creating an easier way to expand available working space. We would also expand the experimental testing to include a more comprehensive analysis of other sampling tasks and integration with other lab equipment such as well plates, centrifuges, and incubators to further enhance the usability of our framework. 

\section{Acknowledgements}

Funding for this work was provided by the Noblis Sponsored Research and Noblis Summer internship programs. The authors declare no conflicts of interest. See Supplementary Table 2 (Table S2)~\cite{logan2025supplemental} for Contributor Role Taxonomy (CRediT) by author and acknowledged contibutors. The following corresponds to author initials. Conceptualization: ZL, AD, DN. Methodology: ZL, AD, DN. Software: ZL. Validation: ZL. Formal analysis: ZL. Investigation: ZL, DN. Resources: ZL, DN. Data Curation: ZL, DN. Writing - Original Draft: ZL, DN. Writing - Review \& Editing: ZL, AD, DN. Visualization: ZL, AD. Supervision: AD, DN. Project administration: AD, DN. Funding acquisition: DN. Conceptualization: KU, GM, NT, BA. Methodology: KU, GM. Investigation: GM. Resources: GM, NT. Writing - Review \& Editing: NT, KJ. Funding acquisition: NT. The authors would like to thank the teams of the Noblis Autonomous Systems Laboratory and the Noblis BSL-II Laboratory for their expertise and resources.


\bibliography{references_zotero}
\bibliographystyle{ieeetr}
\end{document}